\documentclass[9pt,journal,letterpaper,twocolumn]{IEEEtran}
\usepackage{amssymb,amsmath}
\usepackage{balance}
\usepackage{epsfig}
\usepackage{algorithm} 
\usepackage{algorithmic} 
\usepackage{booktabs}
\usepackage{graphicx,subfigure}
\usepackage{graphics}
\usepackage{array}
\usepackage{multirow}
\usepackage{flushend}
\usepackage{color}
\usepackage{amsthm}

\usepackage{ragged2e}
\usepackage{multicol}

\makeatletter

\newcommand{\Rmnum}[1]{\expandafter\@slowromancap\romannumeral #1@}
\makeatother
\ifCLASSOPTIONcompsoc
\usepackage[nocompress]{cite}
\else
\usepackage{cite}
\fi

\ifCLASSINFOpdf
\else
\fi

\begin{document}
	%

	\title{Hierarchical Long Short-Term Concurrent Memory for Human Interaction Recognition}

	%
	%
	%
	%
	
	\author{Xiangbo~Shu,
		Jinhui~Tang,~\IEEEmembership{Senior~Member,~IEEE,}
		Guo-Jun~Qi, Wei~Liu
		and~Jian~Yang
		\thanks{X. Shu, J. Tang and J. Yang are with the School of Computer Science and Engineering, Nanjing
			University of Science and Technology, Nanjing 210094, China (e-mail: shuxb@njust.edu.cn, jinhuitang@njust.edu.cn and csjyang@njust.edu.cn). J. Tang is the corresponding author.}
		\thanks{G.-J. Qi is with the Department of Electrical Engineering and Computer Science, University of Central Florida, Orlando, Florida, 32816, USA (email: guojun.qi@ucf.edu).}
		\thanks{W. Liu is  with  the  Computer  Vision  Group,  Tencent  AI  Lab,  Shenzhen
			518000, China (e-mail: wliu@ee.columbia.edu).}}
	
	%
	%

	\markboth{SUBMISSION~FOR~IEEE~TRANSACTIONS~ON~PATTERN~ANALYSIS~AND~MACHINE~INTELLIGENCE, 2018}%
	{SUBMISSION~FOR~IEEE~TRANSACTIONS~ON~PATTERN~ANALYSIS~AND~MACHINE~INTELLIGENCE, 2018}
	%



	\IEEEcompsoctitleabstractindextext{
		\begin{abstract}
			\justifying
			In this paper, we aim to address the problem of human interaction recognition in videos by exploring the long-term inter-related dynamics among multiple persons. Recently, Long Short-Term Memory (LSTM) has become a popular choice to model individual dynamic for single-person action recognition due to its ability of capturing the temporal motion information in a range. However, existing RNN models focus only on capturing the dynamics of human interaction by simply combining all dynamics of individuals or modeling them as a whole. Such models neglect the inter-related dynamics of how human interactions change over time. To this end, we propose a novel Hierarchical Long Short-Term Concurrent Memory (H-LSTCM) to model the long-term inter-related dynamics among a group of persons for recognizing the human interactions. Specifically, we first feed each person's static features into a Single-Person LSTM to learn the single-person dynamic. Subsequently, the outputs of all Single-Person LSTM units are fed into a novel Concurrent LSTM (Co-LSTM) unit, which mainly consists of multiple sub-memory units, a new cell gate and a new co-memory cell. In a Co-LSTM unit, each sub-memory unit stores individual motion information, while this Co-LSTM unit selectively integrates and stores inter-related motion information between multiple interacting persons from multiple sub-memory units via the cell gate and co-memory cell, respectively. Extensive experiments on four public datasets validate the effectiveness of the proposed H-LSTCM by comparing against baseline  and state-of-the-art methods.
			\justifying
		\end{abstract}
		
		\begin{IEEEkeywords}
			Human interaction recognition, long short-term memory, activity recognition, deep learning.
	\end{IEEEkeywords}}

	\maketitle

	\IEEEdisplaynontitleabstractindextext

	%
	\IEEEpeerreviewmaketitle

	\section{Introduction}

	\IEEEPARstart{H}{uman} interactions (e.g., handshaking, and talking) are typical human activities that occur in public places and are attracting substantial attentions from researchers~\cite{kong2014interactive,chang2015learning,choi2011learning,wang2017a}. A human interaction usually involves at least two individual motions from multiple persons, who are concurrently inter-related with each other (e.g., some persons are talking together, some persons are handshaking with each other). 
	In most cases of human interaction, the concurrent inter-related motions between multiple persons are strongly interacting (e.g., person A kicks person B, while person B retreats back). It has been shown that the concurrent inter-related motions among multiple persons rather than single-person motions can contribute discriminative information for recognizing human interactions~\cite{kong2016close}. 
	

	Two main types of solutions exist for the problem of human interaction recognition. One solution (e.g., \cite{kong2014interactive,kong2012leraning,chang2015learning,zhang2012spatio}) is to extract individual motion descriptors from each interacting person and then predict the class label of an interaction by inferring the coherence between two individual motions. However, this solution, i.e., regarding human interactions as multiple single-person actions, ignores some inter-related motion information and brings in some irrelevant individual motion information. The other solution is to extract motion descriptors on interacting regions and then train an interaction recognition model~\cite{kong2016close}. However, interacting regions are difficult to locate before the close interaction occurs.

	\begin{figure}[t]
		\centering
		\vspace{-0mm}
		\includegraphics[scale=0.275]{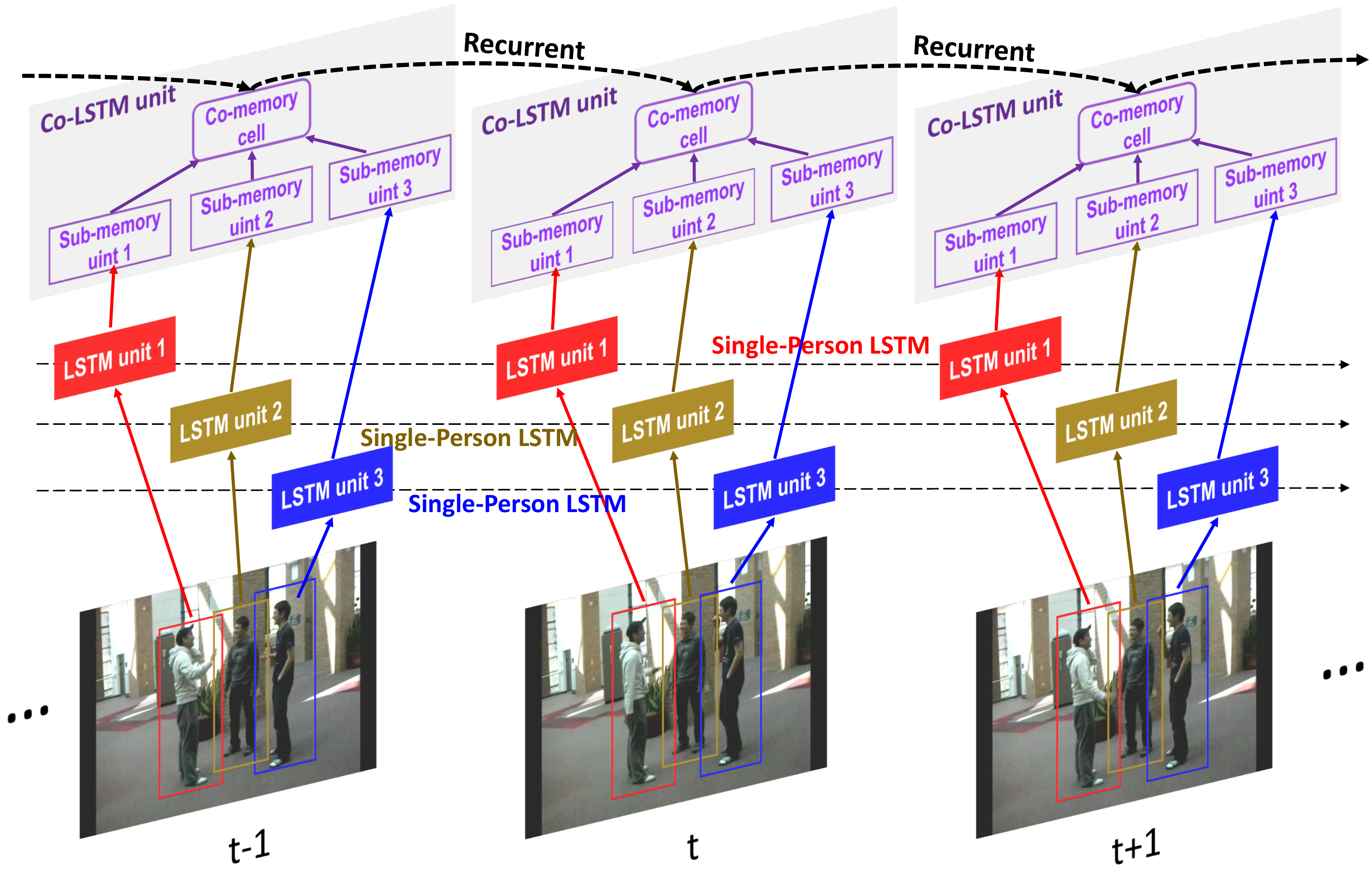}
		\vspace{-4mm}
		\caption{The framework of the proposed Hierarchical Long Short-Term Concurrent Memory (H-LSTCM) for modeling human interactions in a human interaction scene. The details of  Co-LSTM unit is displayed in Figure~\ref{fig_crnn}.}\label{fig_idea}
		\vspace{-3mm}
	\end{figure}
	
	Recently, due to the powerful ability to capture sequential motion information, Long Short-Term Memory (LSTM)~\cite{hochreiter1997long}, has proven to be successful at various human action recognition tasks~\cite{donahue2015long,du2015hierarchical,veeriah2015differential,liu2016spatio,ibrahim2015hierarchical}. Therefore, we aim to explore the long-term inter-related dynamics among a group of interacting persons by leveraging LSTM. However, existing LSTMs model human  dynamics independently, and do not consider the concurrent inter-relation of dynamics among multiple persons. A straightforward way to overcome this limitation is to either 1) merge individual actions at the preprocessing stage~\cite{ke2016spatial} (e.g., consider interacting persons as a whole); or 2) utilize several LSTMs to model the single-person dynamics of individuals and then fuse the output sequences of these LSTMs~\cite{ibrahim2015hierarchical}. 
	However, both methods neglect the inter-related dynamics of how interactions among these persons change over time.

	To this end, we propose a novel Hierarchical Long Short-Term Concurrent Memory (H-LSTCM) for human interaction (activity) recognition to model the long-term inter-related dynamics among a group of interacting persons, as shown in Figure~\ref{fig_idea}. For each person, we first feed her/his static features (e.g., CNN features) into a Single-Person LSTM to learn the single-person dynamic, which describes a person's long-term motion information in a whole video clip. Then, all outputs of Single-Person LSTM units are fed into a novel Concurrent LSTM (Co-LSTM) unit, which mainly consists of multiple sub-memory units, multiple new cell gates and a new co-memory cell. In a Co-LSTM unit, multiple sub-memory units store single-person motion information from the Single-Person LSTM units. Following these sub-memory units, the cell gates allow the inter-related motion memory in sub-memory units to enter a new co-memory cell, and the co-memory cell selectively integrates and stores the inter-related memory to reveal the concurrent inter-related motion information among all interacting persons. Overall, all interacting persons in each frame are jointly modeled by a Co-LSTM unit on the person bonding boxes. At the last time step, the output of Co-LSTM is a dynamic inter-related representation of the group activity. 
	Extensive experiments on various datasets are conducted to evaluate the performance of H-LSTCM compared with the state-of-the-arts.

	The main contributions of this work are summarized as follows:
	\begin{itemize}
		\item We propose a novel Hierarchical Concurrent Long Short-Term Concurrent Memory (H-LSTCM) to effectively address the problem of human interaction recognition with multiple persons, by learning the dynamic inter-related representations among all persons in the group crowd scenes.
		\item We design a novel Concurrent LSTM (Co-LSTM) to aggregate the inter-related memory from individuals in collective activity scenes, by capturing the concurrently long-term inter-related dynamics among multiple persons rather than dynamics of individuals.
	\end{itemize}
	
	{

		Our preliminary Co-LSTSM method in~\cite{shu2017concurrence} with two sub-memory units can recognize only the interactions between two persons, while the proposed H-LSTCM in this paper can recognize various group activities at a larger scale, including collective activity with multiple persons ($\ge 3$ persons), and group activity with multiple sub-group activities. This is because Co-LSTSM learns the dynamic inter-related representation between two persons simply from the static single-person features. Actually there is a large gap between the static single-person features and the dynamic inter-related representation, which limits the performance of the Co-LSTSM. Thus, in H-LSTCM we bring in the single-person dynamic, which is a basic element in the group activity to describe a person's long-term motion information in a whole video clip, and reflects motion patterns caused by interactions with other persons. H-LSTCM learns dynamic inter-related representation among multiple persons in a hierarchical way, from the static to dynamic features at the single-person level first, and further to an inter-related level of group activities. Specifically, the single-person LSTMs in H-LSTCM first learn single-person dynamics from the static single-person features. And then, an extended Co-LSTM with multiple sub-memory units in H-LSTCM learns concurrently inter-related representation among all persons based on the single-person dynamics. Such a hierarchical strategy ensures that H-LSTCM learns more discriminative representation than Co-LSTSM for group activities.

	}
	
	

	\vspace{-2mm}
	\section{Related Work}
	\label{RW}
	
	
	{

		Human interaction recognition (activity recognition) aims to automatically
		understand the interaction performed by at least two persons~\cite{chang2015learning}.
		In the task of two persons' interaction recognition, earlier researchers have noted that several interactive attributes provide discriminative information to represent person-person interactions. For example, Kong et al.~\cite{kong2014interactive,kong2012leraning} regarded multiple interactive phrases as the latent mid-level feature to recognize person-person interactions from human individual actions. Consider that there exists temporal context information in a video clip, Zhang et al.~\cite{zhang2012spatio} and Liu et al.~\cite{liu2011recognizing} used a new set of spatio-temporal action attribute phrases to describe the person-person interactions in a video. However, the difference in some person-person interactions (e.g., boxing and patting) is too small to be identified via only interactive phrases. Moreover, some person-person interactions are complex, and cannot be described well by a specified number of interactive phrases.  
		
		Benefiting from the success of deep learning, some deep learning methods have been proposed to understand two persons' interaction for the last five years~\cite{wang2015hierarchical,ke2016spatial}. For example, Wang {\em et al.}~\cite{wang2015hierarchical} adopted deep context features instead of the traditional context features (e.g.,~\cite{lan2010retrieving}) on the event neighborhood to recognize person-person interactions, where the size of the event neighborhood must be manually defined at the preprocessing step. One limitation of the above methods is that locating the interactive region is a challenging task before the close interaction occurs. Therefore, this work aims to design a human interaction recognition without locating the interactive regions accurately.

		In a scene of multiple persons' interaction (i.e., group activity), several persons interact with each others, which makes activity recognition a complex task.  Two solutions are commonly used to address the problem of group-person interaction recognition. One solution is to exploit spatial distribution of human activities and to present spatio-temporal descriptors to
		capture the spatial distribution of persons~\cite{choi2009they,lan2012discriminative,DBLP:conf/cvpr/RyooA06}.  The other solution  is to  track all the body parts in the video, and then learn holistic representations to estimate the class of the collective activity~\cite{choi2012unified,vahdat2011discriminative}. However, the former solution requires inference of the complex spatio-relation between persons, and the latter brings in some individual action of outlier persons. 

		Recently, Long Short Term Memory (LSTM) has been proposed to address the problem of human interaction recognition by learning high-level dynamic  representations of persons~\cite{deng2015deep,deng2016structure,ke2016spatial}. This insight motivates us to employ superior LSTM models to learn high-level dynamic representations of human activity. Therefore, we propose a new Hierarchical Long Short-Term Concurrent Memory (H-LSTCM) for Human Interaction Recognition. H-LSTCM adopts a hierarchical way to first model the single-person dynamics of individuals by LSTM, and then model the concurrently inter-related dynamics among all the interacting persons by a new Co-LSTM. 
		
		Closely related work includes Hierarchical Deep Temporal Model (HDTM)~\cite{ibrahim2015hierarchical}, Deep Structured Model (DSM)~\cite{deng2015deep}, and Structure Inference Machines (SIM)~\cite{deng2016structure}. Specifically, HDTM~\cite{ibrahim2015hierarchical} first models the individual dynamic motions by several LSTMs. Subsequently, the outputs of these LSTMs are pooled into a single vector, which is the input of a following LSTM. HDTM pools single-person dynamics into an overall dynamic representation, and dose not consider the inter-relations among persons in the group activity. 
		DSM~\cite{deng2015deep} and SIM~\cite{deng2016structure} utilize CNN to obtain the initialized class labels of single-person actions and group-level activity and refine the group activity class label by exploring the relations among the actions of all individuals in an iterative manner. If one person's action is closely related to the group activity and the other persons' actions, this person intensively participates in the group activity; otherwise, this person is an outlier. Since DSM and SIM target ``key" persons who play crucial roles in the group activity rather than those of all persons, some sudden motion information of ``outlier" persons may be lost.”
		Compared to HDTM~\cite{ibrahim2015hierarchical}, the proposed H-LSTCM considers the inter-relations among persons via the cell gates and co-memory cell. And compared to DSM~\cite{deng2015deep} and SIM~\cite{deng2016structure}, the proposed H-LSTCM models the concurrently inter-related dynamics among all persons, which dose lose outlier yet useful persons. 
		
		
		Recently, some works~\cite{alahi2016social,sadeghian2017tracking} proposed to learn the concurrently location-related representation among multiple persons for multi-target tracking. They assume that two persons who have close position are inter-related with each other. By contrast, the proposed H-LSTCM learns sematic-related representation among multiple persons by leveraging the inter-relation between the single-person dynamic at the current time step and the dynamic of the whole activity at the previous time step. Here, it is assumed that one person, whose current representation is closely related to the hidden representation of the whole activity, is likely to be more involved in this activity.
	}
	
	
	\section{Preliminaries: LSTM-Based Action Recognition}
	\label{BK}
	{ Given an input video clip $\{{\bf x}_t\in \mathbb{R}^n|t=1,\cdots,T\}$ with length $T$, where ${\bf x}_t$ is the static feature at time step $t$. A traditional  Recurrent  Neural  Networks (RNN)~\cite{williams1989learning} models the dynamics of this video clip through a sequence of hidden states. Due to the exponential decay in retaining the context information of video frames~\cite{hochreiter1997long}, RNN does not model the long-term dynamics of video sequences well. To this end, Long Short-Term Memory (LSTM)~\cite{hochreiter1997long}, a variant of RNN, provides a solution by incorporating memory units that enable the network to learn when to forget previous hidden states and when to update hidden states given new information~\cite{donahue2015long}. 
		
		
		A traditional  LSTM unit\cite{hochreiter1997long} at time step $t$ contains an input gate ${\bf i}_t$, a forget gate ${\bf f}_t$, an output gate ${\bf o}_t$) and a memory cell ${\bf c }_t$, which are expressed as follows,
		\begin{equation} \label{eq4}
			\begin{aligned}
				{\bf{i}}_t = \sigma ({\bf{W}}_{ix} \cdot {\bf{x}}_t + {\bf{W}}_{ih} \cdot {{\bf{h}}_{t - 1}} + {\bf{b}}_i);
			\end{aligned}
		\end{equation}
		\begin{equation} \label{eq5}
			\begin{aligned}
				{\bf{f}}_t = \sigma ({\bf{W}}_{fx} \cdot {\bf{x}}_t + {\bf{W}}_{fh} \cdot {{\bf{h}}_{t - 1}} + {\bf{b}}_f);
			\end{aligned}
		\end{equation}
		\begin{equation} \label{eq6}
			\begin{aligned}
				{{\bf{o}}_t} = \sigma ({\bf{W}}_{o{{x}}}^{} \cdot{\bf{x}}_t  + {\bf{W}}_{oh} \cdot {{\bf{h}}_{t - 1}} + {\bf{b}}_o),
			\end{aligned}
		\end{equation}
		\begin{equation} \label{eq8}
			\begin{aligned}
				{\bf{g}}_t = \varphi ({\bf{W}}_{gx} \cdot {\bf{x}}_t + {\bf{W}}_{gh} \cdot {{\bf{h}}_{t - 1}} + {\bf{b}}_g),
			\end{aligned}
		\end{equation}
		\begin{equation} \label{eq7}
			\begin{aligned}
				{\bf{c}}_t = {\bf{f}}_t^s \odot {\bf{c}}_{t - 1} + {\bf{i}}_t \odot {\bf{g}}_t,
			\end{aligned}
		\end{equation}
		where $\sigma(\cdot)$ is a sigmoid function, $\odot$ denotes element-wise product, $\varphi(\cdot)$ is a hyperbolic tangent $tanh(\cdot)$, ${\bf{W}}_{*x}$ and ${\bf{W}}_{*h}$ are weight matrices, and ${\bf b}_*$ is bias vector.
		Subsequently, a hidden state ${{\bf{h}}_t}$ at time step $t$ can be expressed as
		\begin{equation} \label{eq9}
			\begin{aligned}
				{{\bf{h}}_t} = {{\bf{o}}_t} \odot \varphi ({{\bf{c }}_t}),
			\end{aligned}
		\end{equation}
		which denotes the dynamic representation of the $t$-th frame. All hidden states $\{{\bf{h}}_t| t=1,2,\cdots,T\}$ describe the dynamic of the video clip.
		Finally, the output ${\bf z}_t\in \mathbb{R}^k$ at time step $t$ is computed as
		\begin{equation} \label{eq2.1}
			\begin{aligned}
				{\bf{z}}_t = \varphi ({\bf{W}}_{zh} \cdot {\bf{h}}_t  + {\bf{b}}_z),
			\end{aligned}
		\end{equation}
		which  can be transformed to a probability $y_{t,l}$ ($l=1,\cdots,k$) corresponding to the $l$-th class of the activity by a softmax function
		\begin{equation} \label{eq3}
			\begin{aligned}
				{{{y}}_{t,l}} = \frac{{\exp ({z_{t,l}})}}{{\sum\limits_{j = 1}^k {\exp ({z_{t,j}})} }},
			\end{aligned}
		\end{equation}
		where $z_{t,j}$ in ${\bf z}_t$ denotes the encoding of the confidence score on the $j$-th activity class. Generally, we set ${{{\bf y}}_{t}}=[y_{t,1},y_{t,2},\cdots,y_{t,k}]^T$ as the predicted class label vector.
	}
	

	\section{Hierarchical Long Short-Term Concurrent Memory}
	\label{PRO}
	\subsection{The Architecture}
	For human interaction recognition, each video frame
	contains at least two concurrent singe-person actions among multiple persons, which are inter-related in a group activity. Existing LSTM models targeting single-person actions cannot handle multiple-person interactions well. As mentioned previously, we can roughly treat all the interacting persons as a whole before training the LSTM network. However, this solution results in some individual-specific motion information. Additionally, we can model the single-person dynamics of individuals by multiple LSTM networks, and then combine (e.g., concatenate or pool) the single-person dynamics obtained by all these LSTM networks into the final representation. Since this strategy assumes that all persons in a group activity are independent of each other, some of the inter-related motion information among these persons is lost.
	\begin{figure}[t]
		\centering
		\includegraphics[scale=0.30]{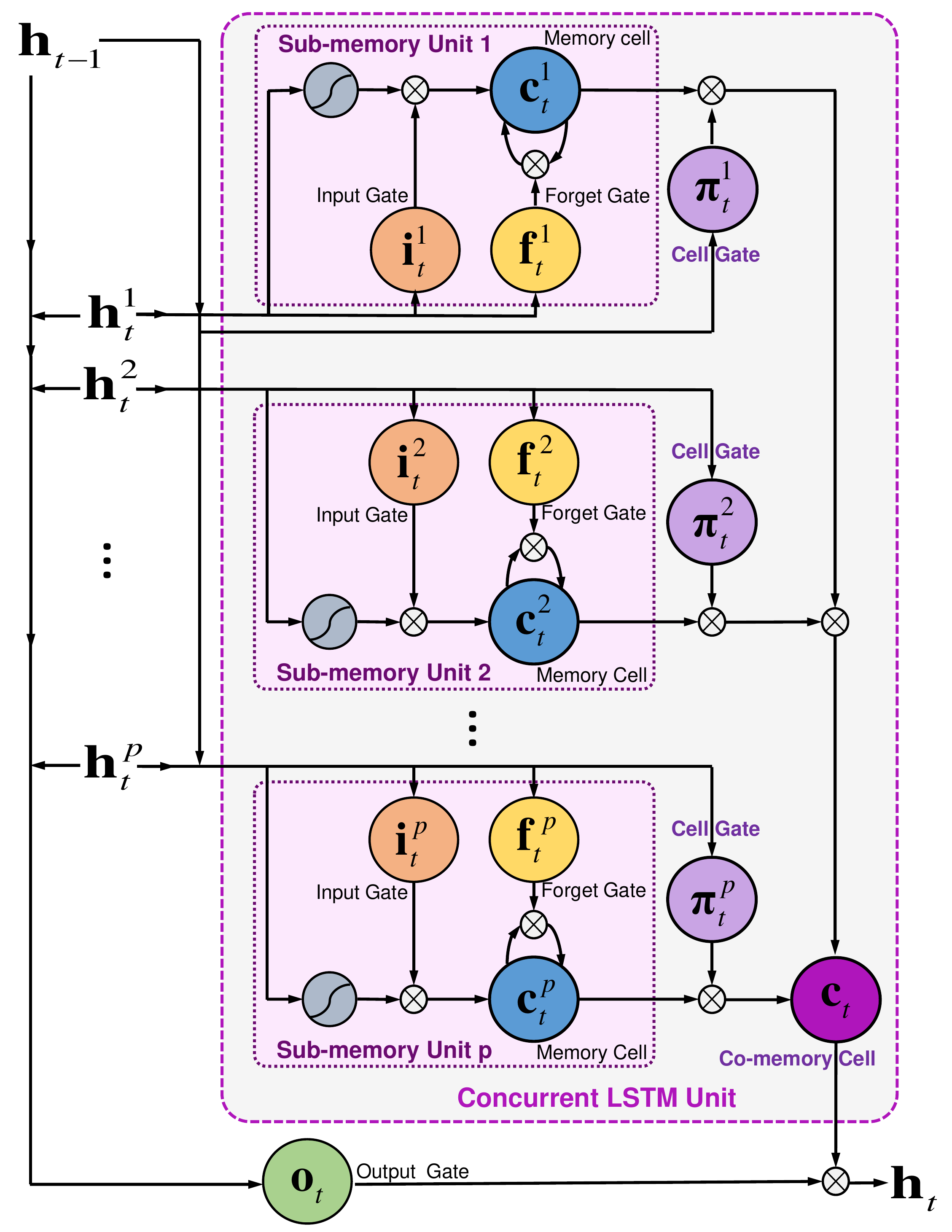}
		\vspace{-1mm}
		\caption{Illustration of a Concurrent LSTM (Co-LSTM) unit in H-LSTCM.}
		\label{fig_crnn}
		\vspace{-3mm}
	\end{figure}
	
	{

		Recently, Deng {\em et al.}~\cite{deng2016structure} proposed a new Structure Inference Machines (SIM) for group activity recognition, which indicated that some persons are related to the group activity, while others are outliers. Specifically, SIM first utilizes CNN to initialize the class labels of individuals' actions and group activity. Then, the group activity class label is refined by considering the relations among the actions of all individuals in an iterative manner. Motivated by this, for a group activity with multiple persons, we also consider designing a model to capture the inter-relation among multiple persons 
		
		However, SIM targets ``key" persons who play crucial roles in the group activity, some sudden motion information of ``outlier" persons may be lost. Empirically, we observe that “outlier” persons are not irrelevant to the group activity at any time step. For example, a person suddenly spike ball at one moment in a volleyball game. Therefore, we propose a Hierarchical Long Short-Term Concurrent Memory (H-LSTCM) to capture the concurrently inter-related dynamics among all the persons rather than those of a selection of persons. Specifically, the proposed H-LSTCM first models the temporal motion information for each person via multiple Single-Person LSTMs corresponding to these persons, and then captures the inter-related dynamics among all the persons by a novel Concurrent LSTM (Co-LSTM). Figure~\ref{fig_idea} shows the whole framework of the proposed H-LSTCM. The key point of H-LSTCM is to utilize multiple sub-memory units in a Concurrent LSTM (Co-LSTM) unit to selectively integrate and store the concurrently inter-related temporal information among multiple persons from the individual temporal information. }

	Figure~\ref{fig_crnn} illustrates the architecture of a Co-LSTM unit of the proposed H-LSTCM at a time step. The Co-LSTM unit mainly consists of multiple specific sub-memory units (the number of units corresponds to the number of interacting persons), multiple cell gates, a common output gate and a new co-memory cell. Specifically, all sub-memory units include their respective input gates, forget gates, and memory cells. Following these sub-memory units, the cell gates allow the inter-related motion memory in the sub-memory units to enter a new co-memory cell, and the co-memory cell selectively integrates and memorizes the inter-related motion information among all the interacting persons. Overall, the stacked Co-LSTM units are recurrent in a time sequence to capture the concurrently inter-related dynamics among all interacting persons over time.

	Formally,  $\{{\bf x}_t^1\in \mathbb{R}^n|t=1,\cdots,T\}$, $\{{\bf x}_t^2\in \mathbb{R}^n|t=1,\cdots,T\}$, $\cdots$ and  $\{{\bf x}_t^p\in \mathbb{R}^n|t=1,\cdots,T\}$ denote the sets of static features (e.g., CNN features) on the tracklets (obtained by object detector and object tracker) of $p$ interacting persons in a video clip $v_n$ (wherein $n=1,2,\cdots,N$). For a feature set of  $\{{\bf x}_t^s|t=1,\cdots,T\}$ of the $s$-th person, we can obtain her/his hidden state (i.e., single-person dynamic) $\{{\bf h}_t^s |t=1,\cdots,T\}$ at each time step via a Single-Person LSTM. In the $s$-th sub-memory unit of Co-LSTM on the top of the Single-Person LSTMs, ${\bf{i}}_t^s$, ${\bf{f}}_t^s$, ${\bf{g}}_t^s$, and  ${\bf{c}}_t^s$ ($s=1,2,\cdots, p$) denote input gate, forget gate, input modulation gate and sub-memory cell at time step $t$, respectively. These components can be expressed by the following equations
	\begin{equation} \label{eq10}
		\begin{aligned}
			{\bf{i}}_t^s = \sigma ({\bf{W}}_{ix}^s \cdot {\bf{h}}_t^s + {\bf{W}}_{ih}^s \cdot {{\bf{h}}_{t - 1}} + {\bf{b}}_i^s), s\!=\! 1,2,\cdots,p;
		\end{aligned}
	\end{equation}
	\begin{equation} \label{eq11}
		\begin{aligned}
			\!\!{\bf{f}}_t^s = \sigma ({\bf{W}}_{fx}^s \cdot {\bf{h}}_t^s + {\bf{W}}_{fh}^s \cdot {{\bf{h}}_{t - 1}} + {\bf{b}}_f^s), s\!=\! 1,2,\cdots,p;
		\end{aligned}
	\end{equation}
	\begin{equation} \label{eq12}
		\begin{aligned}
			\!\!{\bf{g}}_t^s = \varphi ({\bf{W}}_{gx}^s \cdot {\bf{h}}_t^s + {\bf{W}}_{gh}^s \cdot {{\bf{h}}_{t - 1}} + {\bf{b}}_g^s), s\!=\! 1,2,\cdots,p;
		\end{aligned}
	\end{equation}
	\begin{equation} \label{eq13}
		\begin{aligned}
			{\bf{c}}_t^s = {\bf{f}}_t^s \odot {\bf{c}}_{t - 1}^s + {\bf{i}}_t^s \odot {\bf{g}}_t^s, s\!=\! 1,2,\cdots,p,
		\end{aligned}
	\end{equation}
	where $W_{*x}^s$ and $W_{*h}^s$ are weight matrices, $b_*$ is bias vector, and the hidden state ${\bf h}_{t-1}$  denotes the dynamic inter-related representation of the whole activity at time step $t-1$. All hidden states $\{{\bf{h}}_{t}| t=1,2,\cdots,T\}$ describe the inter-related dynamic of the activity scene in the video clip.
	
	Following the $s$-th sub-memory unit, a new cell gate ${\bf \pi}_t^s$ aims to control the memory that enters and leaves this sub-memory unit at time step $t$. Like traditional gates, the cell gate  ${\bf \pi}_t^s$ is activated by a nonlinear function of the input ${\bf h}_t^s$ and the past hidden state ${\bf h}_{t-1}$,
	\begin{equation} \label{eq13.5}
		{{\bf{\pi}}_t^s} = \sigma ({\bf{W}}_{\pi{\rm{h}}}^{s} \cdot 
		{{\bf{h}}_t^s}  + {\bf{W}}_{\pi h}^{} \cdot {{\bf{h}}_{t - 1}} + {\bf{b}}_\pi^{}), s\in \{1,2,\cdots,p\},
	\end{equation}
	where ${\bf W}_{\pi h}^s$ and ${\bf W}_{\pi h}$ are the weight matrices, and ${\bf b}_{\pi}$ is the bias vector.
	Based on the consistent interactions  among multiple interacting persons, all cell gates ${\bf \pi}_t^s$ ($s= 1,2,\cdots,p$) allow more concurrently inter-related motion information among interacting persons to enter a new co-memory cell ${\bf c}_t$, which contributes to a common hidden state ${\bf h}_t$ at time step $t$. In this work, the co-memory cell ${\bf c}_t$ can be expressed as 
	\begin{equation} \label{eq14}
		{{\bf{c}}_t} = \sum\limits_{s = 1}^p {{\bf{\pi }}_t^s \odot {\bf{c}}_t^s}.
	\end{equation}
	This co-memory cell ${\bf c}_t$ corresponds to an output gate ${\bf o}_{t}$ that is related to all the inputs and the common hidden state at the previous time step, i.e.,
	\begin{equation} \label{eq15}
		{{\bf{o}}_t} = \sigma (\sum\limits_{s = 1}^p {{\bf{W}}_{o{\rm{x}}}^s{\bf{h}}_t^s}  + {\bf{W}}_{oh}^{}\cdot{{\bf{h}}_{t - 1}} + {\bf{b}}_o^{}).
	\end{equation}
	Finally, the hidden state ${{\bf{h}}_t}$ at time step $t$ can be expressed as
	\begin{equation} \label{eq16}
		\begin{aligned}
			{{\bf{h}}_t} = {{\bf{o}}_t} \odot \varphi ({{\bf{c }}_t}).
		\end{aligned}
	\end{equation}

	If we obtain ${{\bf{h}}_t}$,  we can compute the probability vector ${\bf y}_t$ of one human interaction by Eq~\eqref{eq2.1} and Eq~\eqref{eq3}.
	
	
	\subsection{Learning Algorithm}

	We employ a loss function to learn the model parameters of H-LSTCM by measuring the deviation between the ground-truth class label vector ${\hat  {\bf y}_t} = [{\hat y}_{t,1},{\hat y}_{t,2},\cdots,{\hat y}_{t,k}]^T$ and the predicted probability vector ${\bf y}_t\!=\![y_{t,l},y_{t,2},\cdots,y_{t,k}]^T$ corresponding to ${\bf h}_t$ at time step $t$,
	\begin{equation} \label{eq17}
		\ell({{\bf{y}}_t},l) =  - \sum\limits_{l = 1}^k {{{\hat y}_{t,l}}{\rm{log}}{y_{t,l}}}.
	\end{equation}
	When the training label of the activity frame at time step $t$ corresponds to the target class ${l_t}$ ($l_t\in\{1,2,\cdots,k\}$), element ${\hat y}_{t,l_t}$ in ${\hat  {\bf y}_t}$ is ${\hat y}_{t,l_t}=1$, and the other elements in ${\hat  {\bf y}_t}$ are zero. Then, Eq.~\eqref{eq17} can be simplified as
	\begin{equation} \label{eq17.1}
		\ell ({{\bf{y}}_t},{l_t}){{ = }} -  \\{\text {log}} {{y}_{t,l_t}},
	\end{equation}
	where ${y}_{t,l_t}$ is defined in Eq.~\eqref{eq3}. Some researchers ~\cite{veeriah2015differential,hochreiter1997long} indicated that the memory cell of LSTM at the last time step can store useful sequence information of the whole data sequence (e.g., a video clip). That is, for a video clip of length $T$, if its class label $l$ is annotated at the video level, the H-LSTCM model can be trained by minimizing the loss at time step $T$, i.e., $\ell ({{\bf{y}}_T},{l}){{ = }} - \log {y_{T,{{l}}}}$. Otherwise, if the class label $l$ is annotated on each frame $t$, we can minimize the cumulative loss over the sequence, i.e., $\sum\nolimits_{t = 1}^T {\ell ({{\bf{y}}_t},{l})}$. 
	
	In this work, given a training video clip with label ${l}$ ($l\in\{1,2,\cdots,k\}$) at the video level, we choose the loss
	\begin{equation} \label{eq19} 
		{\mathcal J }(\Theta)=\ell({{\bf{y}}_T},{l}),
	\end{equation}
	where $\Theta$ denotes a parameter set including all the parameters of the H-LSTCM model. The loss function of H-LSTCM can be minimized by Backpropagation Through Time (BPTT).
	The detailed deductions of the derivatives of all the parameters in the H-LSTCM model can be found in Appendix A of the supplemental material. The detailed training procedure of H-LSTCM is summarized in Algorithm~\ref{alg1}.
	
	\begin{algorithm}[t]
		\scriptsize{
			\renewcommand{\algorithmicrequire}{\textbf{Input:}}
			\renewcommand\algorithmicensure {\textbf{Output:} }
			\caption{Training for H-LSTCM}
			\small
			\label{alg1} 
			\begin{algorithmic}[1]
				\REQUIRE  
				{ $N$ video clips}, $Epoch$, Configuration set of H-LSTCM.\\
				\renewcommand{\algorithmicrequire}{\textbf{Initialization:}} 
				\REQUIRE 
				Parameter set $\Theta$, 
				$epoch\leftarrow1$.
				\STATE Extract fc6 features of each person on the detected bounding box in each frame of each video.\\
				{\em // Forward propagation}
				\STATE Forward propagation of Single-person LSTMs;
				\STATE Forward propagation of Co-LSTM. \\
				{\em // Back propagation}
				\FOR{$epoch = 1,2,\cdots, Epoch$}
				\FOR{$n = 1,2,\cdots, N$}
				\STATE Update parameters in Single-Person LSTMs via BPTT;
				\STATE ${\bf b}_{z}\leftarrow \left(m\cdot{\bf b}_{o}^{s}-\eta\cdot{\frac{{\partial {\mathcal J}(\Theta )}}{{\partial {\bf{b}}_{z}}}} \right)$~\footnotemark[1];
				\STATE ${\bf W}_{zh}\leftarrow \left(m\cdot {\bf b}_{o}^{s}-\eta\cdot{\frac{{\partial {\mathcal J}(\Theta )}}{{\partial {\bf{W}}_{zh}}}} \right)$;
				\STATE ${\bf b}_{o}\leftarrow \left(m\cdot{\bf b}_{o}^{s}-\eta\cdot{\frac{{\partial {\mathcal J}(\Theta )}}{{\partial {\bf{b}}_{o}}}} \right)$;
				\STATE ${\bf W}_{oh}\leftarrow \left(m\cdot{\bf W}_{oh}-\eta\cdot{\frac{{\partial {\mathcal J}(\Theta )}}{{\partial {\bf{W}}_{oh}}}} \right) $;
				\STATE ${\bf W}_{ox}^{s}\leftarrow  \left(m\cdot{\bf W}_{ox}^{s}-\eta\cdot{\frac{{\partial {\mathcal J}(\Theta )}}{{\partial {\bf{W}}_{ox}^s}}} \right)$;
								\STATE ${\bf b}_{\pi}\leftarrow  \left(m\cdot{\bf b}_{\pi}-\eta\cdot{\frac{{\partial {\mathcal J}(\Theta )}}{{\partial {\bf{b}}_{\pi}}}} \right)$;
				\STATE ${\bf W}_{\pi h}^{s}\leftarrow  \left(m\cdot{\bf W}_{\pi h}^{s}-\eta\cdot{\frac{{\partial {\mathcal J}(\Theta )}}{{\partial {\bf{W}}_{\pi h}^s}}} \right)$;
				\STATE ${\bf W}_{\pi h}\leftarrow  \left(m\cdot{\bf W}_{\pi h}-\eta\cdot{\frac{{\partial {\mathcal J}(\Theta )}}{{\partial {\bf{W}}_{\pi h}}}} \right)$;
				\STATE ${\bf b}_{*}^{s}\leftarrow \left(m\cdot{\bf b}_{*}^{s}-\eta\cdot{\frac{{\partial {\mathcal J}(\Theta )}}{{\partial {\bf{b}}_{*}^s}}} \right)$, $s=1,\cdots,p$, and $*\in\{i, f, g\}$;
				\STATE ${\bf W}_{*x}^{s}\leftarrow \left(m\cdot{\bf W}_{*x}^{s}-\eta\cdot{\frac{{\partial {\mathcal J}(\Theta )}}{{\partial {\bf{W}}_{*x}^s}}} \right)$;
				\STATE ${\bf W}_{*h}^{s}\leftarrow \left(m\cdot{\bf W}_{*h}^{s}-\eta\cdot{\frac{{\partial {\mathcal J}(\Theta )}}{{\partial {\bf{W}}_{*h}^s}}} \right)$.		
				\ENDFOR	
				\ENDFOR		
				\ENSURE{Parameter set ${\Theta }$.}			
			\end{algorithmic}
		}
		\footnotemark[1]{Here, $m$ and $\theta$ are the momentum parameter and learning rate, respectively. The detailed deductions of the derivative of all the parameters can be found in Appendix A.}
	\end{algorithm}

	
	\section{Experiments}
	\label{EXP}
	In experiments, we evaluate the performance of H-LSTCM compared with the state-of-the-art methods and some baselines on four public datasets. 
	
\begin{table*}[t]
	\caption{Recognition  accuracy (\%) on the BIT dataset.}
	\vspace{-3.5mm}
	{\scriptsize
		\begin{center}
			\begin{tabular}{lccccccccc}
				\hline
				\centering
				Method & bow & boxing & handshake & high-five & hug & kick & pat & push & Average \\
				\hline
				Lan {\em et al.}~\cite{lan2012discriminative} & 81.25 & 75.00 & 81.25 & 87.50 & 87.50 & 81.25 & 81.25 & 81.25 & 82.03 \\
				Liu {\em et al.}~\cite{liu2011recognizing} & 100.00 & 75.00 & 81.25 & 87.50 & 93.75 & 87.50 & 75.00 & 75.00 & 84.37 \\
				Kong {\em et al.}~\cite{kong2012leraning} & 81.25 & 81.25 & 81.25 & 93.75 & 93.75 & 81.25 & 81.25 & 87.50 & 85.16 \\
				Kong {\em et al.}~\cite{kong2016close} & 87.50 & 81.25 & 87.50 & 81.25 & 87.50 & 81.25 & 87.50 & 87.50 & 85.38 \\
				Kong {\em et al.}~\cite{kong2014interactive} & 93.75 & 87.50 & {93.75} & {93.75} & 93.75 & 87.50 & 87.50 & 87.50 & 90.63 \\
				Donahue {\em et al.}~\cite{donahue2015long} & 100.00 & 75.00 & 85.00&	69.75& 85.00& 69.75	&80.00	&76.50 &80.13 \\
				Ke {\em et al.}~\cite{ke2016spatial} & - & - &	-&	-&	-&	-&	-&	- & 85.20 \\
				\hline
				B1    & 100.00 & 75.00& 62.50& 	56.25& 	93.75& 	68.75& 	56.25& 	62.50& 	71.88\\
				B2      & 100.00 & 75.00&	84.50&	84.50&	88.00&	88.00&	70.00&	78.00&	83.50 \\
				B3      & 100.00 & 79.00& 84.50&	84.50&	{94.75}&	88.00&	80.50&	90.00&	87.66
				\\
				B4　& 100.00	& 82.00 &	85.75 &	84.50 & 94.75 &	88.00	& 83.00	& 90.00 & 88.50
				
				\\
				{Co-LSTSM}~\cite{shu2017concurrence}　& {100.00} &	{90.50} &	{92.50} &	{92.50} &	{94.75}&	{88.00} &	{90.50} &	{94.25} &	{92.88}	
				\\
				{H-LSTCM}　& {100.00} &	{92.50} &	{94.75} &	{95.50} &	{94.75}&	{89.50} &	{91.00} &	{94.25} &	{94.03}	
				\\
				\hline
			\end{tabular}
		\end{center}
	}
\vspace{-3mm}
	\label{BIT_results}
\end{table*}
	
	\vspace{-2mm}
	\subsection{Datasets}
	The detailed descriptions of four public datasets are as follows:	
	\begin{itemize}	
		\item	{\bf BIT dataset~\cite{kong2012leraning}.} It consists of eight classes of human interactions, i.e., bow, boxing, handshake, high-five, hug, kick,
		pat, and push. Each class includes 50 videos with cluttered  backgrounds.  Following in~\cite{kong2014interactive}, 34 videos per class are randomly chosen as the training data and the remaining ones are used for testing.	
		\item 	{\bf UT dataset~\cite{ryoo2009spatio}.}
		It consists of  ten  videos, each video containing six classes of human interactions, i.e., handshake, hug, kick, point, punch and push. After extracting the frames, we obtain 60 video clips, namely 10 video clips per class. Leave-one-out cross-validation is adopted for the experiments.
		
		\item 	{\bf Collective Activity Dataset (CAD)~\cite{choi2009they}.} It contains 44 videos of five multiple-person activities, i.e., crossing, waiting, queuing, walking, and talking. Similar to~\cite{lan2012discriminative,hajimirsadeghi2015visual}, we select one-third of the video clips from each activity category to form the test set, and the rest of the video clips are used for training. The one-versus-all technique is employed for this recognition task.		
		
		{ \item {\bf Volleyball Dataset (VD)~\cite{ibrahim2015hierarchical}.} It contains 55 volleyball videos with 4830 annotated frames. Each frame, there has a group-level activity class label (e.g., left\_pass, right\_pass, left\_set,  right\_set, left\_spike, right\_spike, left\_winpoint or right\_winpoint). Following in \cite{ibrahim2015hierarchical}, two-thirds of the annotated frames are used for training and the remaining ones are used for testing.}
	\end{itemize}

	\subsection{Implementation Details}
	In the preprocessing step, the bounding box (tracklet) corresponding to each  person is  detected  and  tracked  over all frames  by
	an object detector~\cite{girshick2015fast} and an object tracker~\cite{zamir2012gmcp}. Following in~\cite{donahue2015long}, the pretrained AlexNet model~\cite{krizhevsky2012imagenet} is employed to extract the fc6 feature (static feature) on each bounding box around one person, respectively.

	For the BIT, UT, CAD and VD datasets, the length $T$ of the time steps is set to $30$, $40$, $10$ and $10$, respectively. 
	In the configurations of H-LSTCM on four datasets, the number of memory cell nodes of each Single-Person LSTM, the number of output nodes of each Single-Person LSTM, and the number of sub-memory cell nodes of Co-LSTM are set to 2048, 1024 and 512, respectively. We use the Torch toolbox and Caffe~\cite{jia2014caffe} as the deep learning platform and an NVIDIA Tesla K20 GPU to run the experiments. The learning rate, momentum and decay rate are set to $0.5\times10^{-3}$, 0.9 and 0.95, respectively. In experiments, the training of H-LSTCM begins to converge after approximately $500$, $600$, $600$ and $700$ epochs on the BIT, UT, CAD and VD datasets, respectively. The learning curves for training the proposed H-LSTCM on the BIT, UT, CAD and VD datasets are plotted in Appendix B of the supplemental material.
	

	In experiments, the following four baselines are chosen:
	\begin{itemize}
		\item {\bf B1: Person-box CNN}. The pre-trained AlexNet is deployed on each person bounding box at each time step, where all fc6 features corresponding to each person are concatenated into a long vector. Then the concatenated features over all time steps are pooled into a single feature. All features from each video clip are trained and tested by the softmax classifier. This baseline illustrates the importance of deep features.
		\item
		{\bf B2: One CNN + LSTM}. This baseline treats two individual actions as a whole.   First, multiple bounding boxes corresponding to each interacting person respectively at each time step are merged into a larger bounding box. Second, fc6 features are extracted by AlexNet on this ``larger" bounding box at each time step. Third, we use the fc6 features as inputs to train an LSTM. This baseline is similar to that of Long-term Recurrent Convolutional Networks~\cite{donahue2015long}.
		\item
		{\bf B3: Multiple CNN + LSTM}.  This  baseline models the individual dynamics of multiple persons by Multiple LSTMs. First, AlexNet is deployed on each person bounding box at each time step to extract the fc6 feature. Second, the fc6 feature extracted from each person is fed into an LSTM network to capture the individual dynamic motions, respectively. Third, we average the softmax scores output of all LSTM networks. Here, the averaged score reflects the probability of the action class. This baseline is the same as Two-Stream Convolutional Networks~\cite{simonyan2014two}.
		\item
		{\bf B4: Single-Person LSTMs + Whole LSTM}. This baseline learns the single-person dynamics via multiple LSTMs, and the outputs are pooled into the other LSTM. Specifically, we first use AlexNet to extract fc6 features on person bounding boxes at each frame. Second, the fc6 features of each person are fed into each traditional LSTM network to learn the single-person hidden states. Third, the hidden states of all persons at each time step are max pooled into a single vector, which is fed into the other LSTM network, followed by a softmax. This baseline is the same as Hierarchical Deep Temporal Models~\cite{ibrahim2015hierarchical}.
		
	\end{itemize}
	

	\subsection{Results on the BIT dataset}
	
	{\bf Comparison with baselines.} Table~\ref{BIT_results} shows the recognition accuracy of the proposed H-LSTCM are better than all baseline methods. Adding the temporal information by employing LSTM (i.e., B2, B3, B4, and Co-LSTSM) improves the performance of B1 without temporal information. Specifically,  Co-LSTSM achieves higher accuracy than B2, B3 and B4. It is illustrated that inter-related motion information among multiple persons is more important than the single-person motion information of individuals for recognizing human interactions. The confusion matrix of H-LSTCM is shown in Appendix C of the supplementary material. 

\begin{figure}[!t]
		\centering
		\includegraphics[scale=0.36]{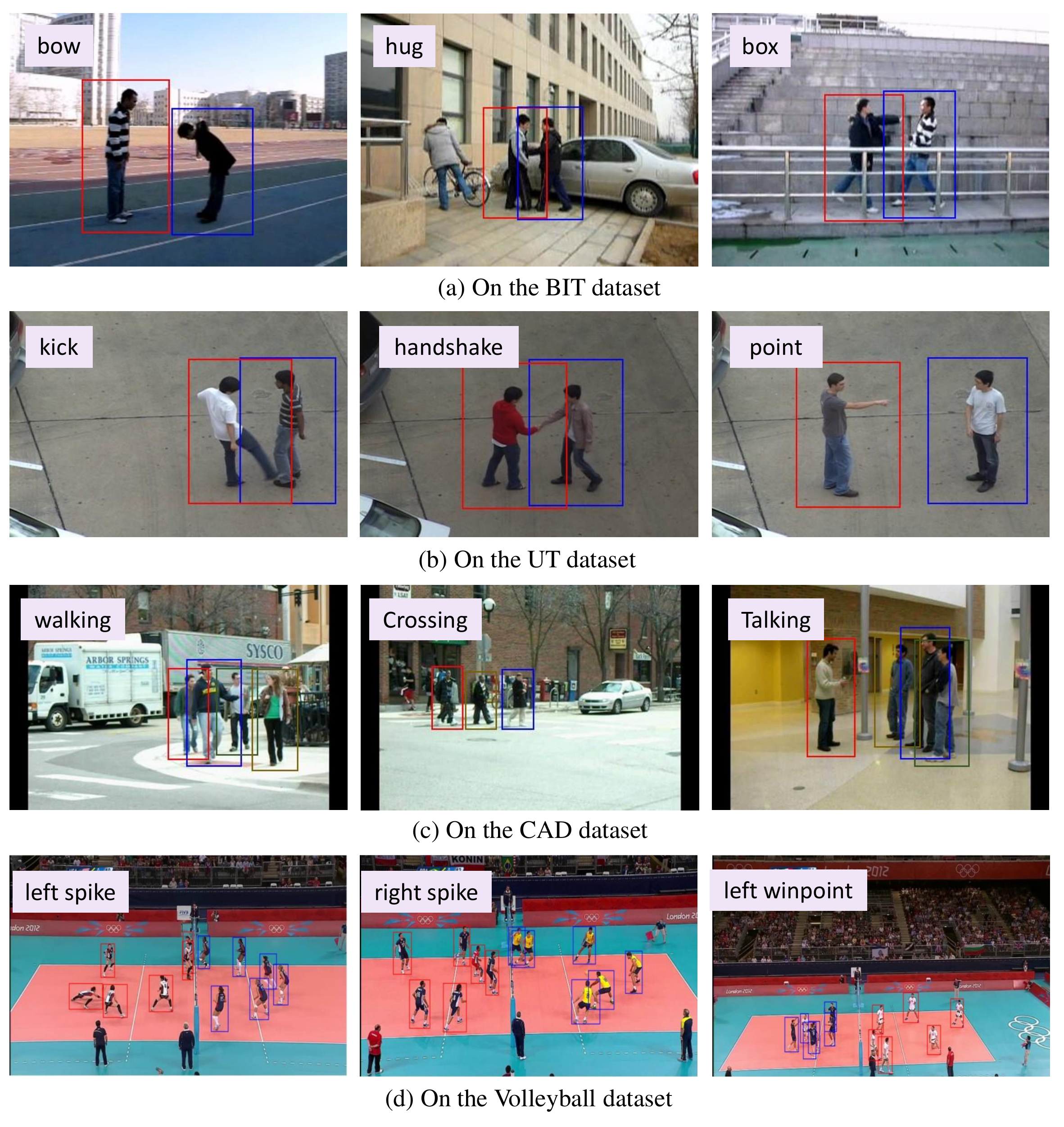}
		\vspace{-2mm}
		\caption{ Examples of some recognition results of the proposed method on four datasets. In the  BIT, UT and CAD datasets, each person is detected and tracked by a bounding box, of which the size is enlarged by a moderate scale to cover more context information. The VD dataset provides person bounding boxes. Better view in color.}\label{fig_results}
		\vspace{-1mm}
	\end{figure}

	{\bf  Comparison with state-of-the-art methods.} We also compare H-LSTCM
	with the state-of-the-art methods for human interaction recognition, i.e., hand-crafted spatio-temporal interest
	points~\cite{DollarVSPETS05cuboids} methods of Lan {\em et al.}~\cite{lan2012discriminative}, Liu {\em et al.}~\cite{liu2011recognizing}, and Kong {\em et al.}~\cite{kong2012leraning,kong2014interactive,kong2016close}, as well as the LSTM-based methods of  Donahue {\em et al.}~\cite{donahue2015long}, and Ke {\em et al.}~\cite{ke2016spatial}. Table~\ref{BIT_results} lists the results of recognition accuracy, in which some results are reported in~\cite{kong2014interactive,kong2016close}. H-LSTCM performs better than the alternatives, especially the LSTM-based methods, i.e., Donahue {\em et al.}~\cite{donahue2015long} and Ke {\em et al.}~\cite{ke2016spatial}. In particular, the proposed H-LSTCM has gained an approximately 9\% improvement compared
	with the state-of-the-art LSTM-based methods (i.e., Ke {\em et al.}~\cite{ke2016spatial} with an accuracy of 85.20\%). Some examples of the recognition results of H-LSTCM are shown in Figure~\ref{fig_results}(a).
	
	\subsection{Results on the UT dataset}
	\vspace{-0mm}
	{\bf Comparison with baselines.} Table~\ref{UT_results} shows the recognition accuracy of the proposed H-LSTCM compared with that of baselines (including Co-LSTSM~\cite{shu2017concurrence}). It is observed that H-LSTCM performs consistently better than all the baselines. In particular, H-LSTCM and Co-LSTSM, targeting to model the inter-related dynamics rather than the individual dynamics, achieve impressive accuracy. The confusion matrix of H-LSTCM is shown in Appendix C of the supplementary material. 
	
	
		\begin{table}[!t]
		\caption{Recognition  accuracy (\%) of different methods on UT dataset.}
		\vspace{-3mm}
		{\scriptsize
			\begin{center}
				\begin{tabular}{lc cc cccc}
					\hline\hline
					\centering
					\hspace{-0.5em}Method & \hspace{-0.5em}handshake\hspace{-0.5em} & \hspace{-0.5em}hug\hspace{-0.5em} & \hspace{-0.5em}kick\hspace{-0.5em} & \hspace{-0.5em}point\hspace{-0.5em} & \hspace{-0.5em}punch\hspace{-0.5em} & \hspace{-0.5em}push\hspace{-0.5em} & \hspace{-0.5em}Average\hspace{-0.5em}  \\
					\hline
					\hspace{-0.5em}Ryoo {\em et al.}~\cite{ryoo2009spatio} & \hspace{-0.5em}75.00\hspace{-0.5em} & \hspace{-0.5em}87.50\hspace{-0.5em} & \hspace{-0.5em}62.50\hspace{-0.5em} & \hspace{-0.5em}50.00\hspace{-0.5em} & \hspace{-0.5em}75.00\hspace{-0.5em} & \hspace{-0.5em}75.00\hspace{-0.5em} & \hspace{-0.5em}70.80\hspace{-0.5em}  \\
					\hspace{-0.5em}Yu {\em et al.}~\cite{yu2010real} & \hspace{-0.5em}~100.00\hspace{-0.5em}& \hspace{-0.5em}65.00\hspace{-0.5em}& \hspace{-0.5em}{100.00}\hspace{-0.5em} & \hspace{-0.5em}85.00\hspace{-0.5em}& \hspace{-0.5em}75.00\hspace{-0.5em}& \hspace{-0.5em}75.00\hspace{-0.5em}& \hspace{-0.5em}83.33\hspace{-0.5em} \\
					\hspace{-0.5em}Ryoo ~\cite{ryoo2011human} & \hspace{-0.5em}80.00\hspace{-0.5em} & \hspace{-0.5em}90.00\hspace{-0.5em} & \hspace{-0.5em}90.00\hspace{-0.5em} & \hspace{-0.5em}80.00\hspace{-0.5em} & \hspace{-0.5em}90.00\hspace{-0.5em} & \hspace{-0.5em}80.00\hspace{-0.5em} & \hspace{-0.5em}85.00\hspace{-0.5em} \\
					\hspace{-0.5em}Kong {\em et al.}~\cite{kong2012leraning} & \hspace{-0.5em}80.00\hspace{-0.5em} & \hspace{-0.5em}80.00\hspace{-0.5em} & \hspace{-0.5em}{100.00}\hspace{-0.5em} & \hspace{-0.5em}90.00\hspace{-0.5em} & \hspace{-0.5em}90.00\hspace{-0.5em} & \hspace{-0.5em}90.00\hspace{-0.5em} & \hspace{-0.5em}88.33\hspace{-0.5em} \\
					\hspace{-0.5em}Kong {\em et al.}~\cite{kong2014interactive}  & \hspace{-0.5em}100.00\hspace{-0.5em} & \hspace{-0.5em}90.00\hspace{-0.5em} & \hspace{-0.5em}{100.00}\hspace{-0.5em} & \hspace{-0.5em}80.00\hspace{-0.5em} & \hspace{-0.5em}90.00\hspace{-0.5em} & \hspace{-0.5em}90.00\hspace{-0.5em} & \hspace{-0.5em}91.67\hspace{-0.5em}  \\
					\hspace{-0.5em}Kong {\em et al.}~\cite{kong2016close} & \hspace{-0.5em}90.00\hspace{-0.5em} & \hspace{-0.5em}100.00\hspace{-0.5em} & \hspace{-0.5em}90.00\hspace{-0.5em} & \hspace{-0.5em}100.00\hspace{-0.5em} & \hspace{-0.5em}90.00\hspace{-0.5em} & \hspace{-0.5em}90.00\hspace{-0.5em} & \hspace{-0.5em}93.33\hspace{-0.5em}  \\
					\hspace{-0.5em}Raptis et al.~\cite{raptis2013poselet} & \hspace{-0.5em}100.00\hspace{-0.5em} & \hspace{-0.5em}100.00\hspace{-0.5em} & \hspace{-0.5em}90.00\hspace{-0.5em} & \hspace{-0.5em}100.00\hspace{-0.5em} & \hspace{-0.5em}80.00\hspace{-0.5em} & \hspace{-0.5em}90.00\hspace{-0.5em} & \hspace{-0.5em}93.30\hspace{-0.5em} \\
					\hspace{-0.5em}Shariat et al.~\cite{shariat2013a} & - & - & - & - & - & - & 91.57 \\
					\hspace{-0.5em}Zhang {\em et al.}~\cite{zhang2012spatio} & \hspace{-0.5em}100.00\hspace{-0.5em} & \hspace{-0.5em}100.00\hspace{-0.5em} & \hspace{-0.5em}{100.00}\hspace{-0.5em}  & \hspace{-0.5em}90.00\hspace{-0.5em} & \hspace{-0.5em}90.00\hspace{-0.5em} & \hspace{-0.5em}90.00\hspace{-0.5em} & \hspace{-0.5em}{95.00}\hspace{-0.5em} \\
					\hspace{-0.5em}Donahue {\em et al.}~\cite{donahue2015long} & \hspace{-0.5em}~90.00\hspace{-0.5em}& \hspace{-0.5em}80.00\hspace{-0.5em} &\hspace{-0.5em}90.00\hspace{-0.5em}	& \hspace{-0.5em}80.00\hspace{-0.5em} & \hspace{-0.5em}90.00\hspace{-0.5em}	& \hspace{-0.5em}~80.00\hspace{-0.5em}&  \hspace{-0.5em}85.00\hspace{-0.5em} \\
					\hspace{-0.5em}Ke {\em et al.}~\cite{ke2016spatial}  &	-&	-&	-&	-&	-&	- & 93.33\\
					\hspace{-0.5em}Wang {\em et al.}~\cite{wang2015hierarchical}  &	-&	-&	-&	-&	-&	- & \hspace{-0.5em}{95.00}\hspace{-0.5em} \\
					\hline
					\hspace{-0.5em}B1  & \hspace{-0.5em}90.00\hspace{-0.5em} & \hspace{-0.5em}80.00\hspace{-0.5em} & \hspace{-0.5em}80.00\hspace{-0.5em} & \hspace{-0.5em}80.00\hspace{-0.5em} & \hspace{-0.5em}80.00\hspace{-0.5em} & \hspace{-0.5em}80.00\hspace{-0.5em} & \hspace{-0.5em}81.67\hspace{-0.5em}   \\
					\hspace{-0.5em}B2  & \hspace{-0.5em}90.00\hspace{-0.5em}	& \hspace{-0.5em}80.00\hspace{-0.5em}	& \hspace{-0.5em}90.00\hspace{-0.5em}	& \hspace{-0.5em}80.00\hspace{-0.5em}	& \hspace{-0.5em}90.00\hspace{-0.5em}	& \hspace{-0.5em}80.00\hspace{-0.5em} & \hspace{-0.5em}~85.00\hspace{-0.5em}\\
					\hspace{-0.5em}B3 & \hspace{-0.5em}100.00\hspace{-0.5em}	& \hspace{-0.5em}100.00\hspace{-0.5em}	& \hspace{-0.5em}90.00\hspace{-0.5em}	& \hspace{-0.5em}80.00\hspace{-0.5em}	& \hspace{-0.5em}90.00\hspace{-0.5em}	& \hspace{-0.5em}80.00\hspace{-0.5em} & \hspace{-0.5em}~90.00\hspace{-0.5em}\\
					\hspace{-0.5em}B4 & \hspace{-0.5em}{100.00}\hspace{-0.5em} &	\hspace{-0.5em}{100.00}\hspace{-0.5em} & \hspace{-0.5em}{90.00}\hspace{-0.5em}	& \hspace{-0.5em}{90.00}\hspace{-0.5em} & \hspace{-0.5em}{90.00}\hspace{-0.5em}	& \hspace{-0.5em}{80.00}\hspace{-0.5em} & \hspace{-0.5em}{91.67}\hspace{-0.5em}  \\
					\hspace{-0.5em}{Co-LSTSM}~\cite{shu2017concurrence} & \hspace{-0.5em}{100.00}\hspace{-0.5em} &	\hspace{-0.5em}{100.00}\hspace{-0.5em} & \hspace{-0.5em}{90.00}\hspace{-0.5em}	& \hspace{-0.5em}{100.00}\hspace{-0.5em} & \hspace{-0.5em}{90.00}\hspace{-0.5em}	& \hspace{-0.5em}{90.00}\hspace{-0.5em} & \hspace{-0.5em}{95.00}\hspace{-0.5em}  \\
					\hspace{-0.5em}{H-LSTCM} & \hspace{-0.5em}{100.00}\hspace{-0.5em} &	\hspace{-0.5em}{100.00}\hspace{-0.5em} & \hspace{-0.5em}{100.00}\hspace{-0.5em}	& \hspace{-0.5em}{100.00}\hspace{-0.5em} & \hspace{-0.5em}{100.00}\hspace{-0.5em}	& \hspace{-0.5em}{90.00}\hspace{-0.5em} & \hspace{-0.5em}{98.33}\hspace{-0.5em}  \\
					\hline\hline
				\end{tabular}
			\end{center}
		}
		\label{UT_results}	
		\vspace{-3.5mm}		
	\end{table}

	{\bf Comparison with state-of-the-art methods.} The proposed H-LSTCM is also compared
	with the state-of-the-art methods, including some traditional methods (i.e., Ryoo {\em et al.}~\cite{ryoo2009spatio}, Yu {\em et al.}~\cite{yu2010real}, Kong {\em et al.}~\cite{kong2012leraning,kong2014interactive,kong2016close},
	Raptis {\em et al.}~\cite{raptis2013poselet}, Shariat {\em et al.}~\cite{shariat2013a}, and Zhang {\em et al.}~\cite{zhang2012spatio}), a deep learning method (i.e., Wang {\em et al.}~\cite{wang2015hierarchical}), as well as LSTM-based methods (i.e., Ke {\em et al.}~\cite{ke2016spatial} and Donahue {\em et al.}~\cite{donahue2015long}).
	The recognition accuracy results are shown in Table~\ref{UT_results}. Co-LSTSM achieves satisfactory accuracy, i.e., 95\%. By further extending Co-LSTSM in a hierarchical way, the proposed H-LSTCM, which first models single-person dynamics and then captures concurrently inter-related dynamics among persons, improves the recognition accuracy to $98.33\%$, which is the state-of-the-art performance. Some of the recognition results of H-LSTCM are shown in Figure~\ref{fig_results}(b).

	\subsection{Results on the CAD dataset}	
	\label{cad}	
	{\bf Comparison with baselines.} We compare the recognition accuracy of the proposed H-LSTCM and that of all the baselines. We also regard the preliminary Co-LSTSM~\cite{shu2017concurrence} as a baseline. Since most of the group activities in the CAD dataset contain multiple interacting persons ($\ge 3$ persons), the original Co-LSTSM~\cite{shu2017concurrence} modeling two interacting persons cannot directly model group activity with multiple persons ($\ge 3$ persons). Thus, we extend the Co-LSTSM to a new version, named as Co-LSTSM$^+$. Co-LSTSM$^+$ has multiple sub-memory units corresponding to multiple persons, and its architecture is similar to the Co-LSTM module of H-LSTCM in Figure 3. The recognition accuracy of the proposed H-LSTCM and all baselines is shown in Table~\ref{CAD_results}. It is observed that H-LSTCM achieves the best performance. Furthermore, Co-LSTSM+ no longer achieves the significant performance improvements compared with B4. Here, Co-LSTSM$^+$ (i.e., the extended version of Co-LSTSM~\cite{shu2017concurrence}) cannot capture the complex inter-related dynamics among multiple persons based on the static single-person CNN features, since the collective activities in the CAD dataset are more complex than the interactions in either the BIT dataset or UT dataset. The confusion matrix of H-LSTCM is shown in Appendix C of the supplementary material.    
	
		\begin{table}[!t]
			\vspace{-2mm}
		\caption{{Recognition accuracy (\%) of different methods on CAD.}}
		\vspace{-3mm}
		{\scriptsize
			\begin{center}
				\begin{tabular}{lc cc ccc}
					\hline
					\centering
					\hspace{-0.5em}Method & \hspace{-0.5em}crossing\hspace{-0.5em} &\hspace{-0.5em} waiting\hspace{-0.5em} & \hspace{-0.5em}queuing\hspace{-0.5em} & \hspace{-0.5em}walking\hspace{-0.5em} & \hspace{-0.5em}talking\hspace{-0.5em} & \hspace{-0.5em}Average\hspace{-0.5em}  \\
					\hline
					\hspace{-0.5em}Choi {\em et al.}~\cite{choi2009they}\hspace{-0.5em} & \hspace{-0.5em}55.4\hspace{-0.5em} & \hspace{-0.5em}64.6\hspace{-0.5em} & \hspace{-0.5em}63.3\hspace{-0.5em} & \hspace{-0.5em}57.9\hspace{-0.5em}  & \hspace{-0.5em}83.6\hspace{-0.5em} & \hspace{-0.5em}65.9\hspace{-0.5em} \\
					\hspace{-0.5em}Lan {\em et al.}~\cite{lan2010retrieving}\hspace{-0.5em}& \hspace{-0.5em}75\hspace{-0.5em} & \hspace{-0.5em}74\hspace{-0.5em} & \hspace{-0.5em}74\hspace{-0.5em}& \hspace{-0.5em}57 &61  & 68.2  \\
					\hspace{-0.5em}Choi {\em et al.}~\cite{choi2011learning}& 76.4  & 76.4 & 78.7 & 36.8 & 85.7 & 70.9  \\
					\hspace{-0.5em}Antic {\em et al.}~\cite{antic2014learning} & 73.70 & 74.50 & 90.10 & 62.00 & 70.00 & 74.1 \\
					\hspace{-0.5em}Liu {\em et al.}~\cite{liu2011recognizing}& 72.73  & 66.67  & 71.43  & 83.33 & 85.71 & 76.19  \\
					\hspace{-0.5em}Wang  {\em et al.}~\cite{wang2017a} & 64.8 & 66.0 & 66.7 & 89.2 & 99.5  & 77.2 \\	
					\hspace{-0.5em}Lan {\em et al.}~\cite{lan2012discriminative} & 68 & 69 & 76 & 80  &  99& 79.7  \\				
					\hspace{-0.5em}Choi {\em et al.}~\cite{choi2012unified} &  61.3 & 82.9 & 95.4 & 65.1 & 94.9 & 79.9   \\		
					\hspace{-0.5em}Kong {\em et al.}~\cite{kong2014interactive}  & 77.27  & 77.78 & 85.71 & 83.33 & 100 & 82.54   \\
					\hspace{-0.5em}Zhou {\em et al.}~\cite{zhou2016generative} &  76.83 & 74.36 & 93.76 & 87.63 & 98.16 & 82.07   \\			    
					\hspace{-0.5em}Ibrahim {\em et al.}~\cite{ibrahim2015hierarchical} & 61.54 & 66.44  & 96.77 & 80.41 & 99.45 & 81.50   \\
					\hspace{-0.5em}{\scriptsize \scriptsize Hajimirsadeghi {\em et al.}}~\cite{hajimirsadeghi2016multi-instance}  & 72 & 75   & 92 & 70 & 99 & 81.6 \\		
					\hspace{-0.5em}Deng {\em et al.}~\cite{deng2015deep}  &  - & - & - & - & - & 80.6   \\
					\hspace{-0.5em}Deng {\em et al.}~\cite{deng2016structure} &  - & - & - & - & - & 81.2  \\
					
					\hline
					B1   & 46.21 &	53.69 &	70.20&	61.19&	74.33&	61.12
					\\
					B2  & 52.38	& 54.50&	73.89&	61.45&	76.35&	63.71
					\\
					B3 & 52.46&	54.61&	82.00&	61.21&	79.85&	66.02
					\\
					B4 & 62.60& 	65.25& 	90.74& 	78.33& 	95.36& 	78.46
					\\
					{Co-LSTSM}$^{+}$  & 65.50&	64.85&	94.67&	75.33&	95.33&
					79.14
					\\
					{H-LSTCM} & 65.50&	68.29&	97.90&	87.69&	99.35&	83.75
					\\
					\hline
				\end{tabular}
			\end{center}
		}
		\vspace{-3mm}
		\label{CAD_results}	
	\end{table}
	

	{\bf Comparison with state-of-the-art methods.} We also compare the recognition accuracy of H-LSTCM and the state-of-the-art methods, including some traditional methods (i.e., Choi {\em et al.}~\cite{choi2009they}, Lan {\em et al.}~\cite{lan2010retrieving}, Choi {\em et al.}~\cite{choi2011learning}, Antic {\em et al.}~\cite{antic2014learning}, Liu {\em et al.}~\cite{liu2011recognizing}, Wang  {\em et al.}~\cite{wang2017a}, Lan {\em et al.}~\cite{lan2012discriminative}, Choi {\em et al.}~\cite{choi2012unified}, Kong {\em et al.}~\cite{kong2014interactive}, Zhou {\em et al.}~\cite{zhou2016generative}, and Hajimirsadeghi {\em et al.}~\cite{hajimirsadeghi2016multi-instance}), a deep learning based method (i.e., Deng {\em et al.}~\cite{deng2015deep}), RNN based methods (i.e., Deng {\em et al.}~\cite{deng2016structure,ibrahim2015hierarchical}, and an LSTM based method (i.e., Ibrahim {\em et al.}~\cite{ibrahim2015hierarchical}).
	The results of recognition accuracy are shown in Table~\ref{UT_results}. H-LSTCM achieves better performance than that of the other methods. As a new exploration that leverages the variants of LSTM, the proposed H-LSTCM achieves an approximately $2\%$ improvement compared with the most closely related work~\cite{ibrahim2015hierarchical}, which uses only the traditional LSTM model without any change. Finally, we present some recognition results of H-LSTCM in Figure~\ref{fig_results}(c).


	{ \subsection{Results on the Volleyball dataset}		
		{\bf Comparison with baselines.} In a volleyball sport, there are two sub-groups of players from two teams. The players on the same team have more interactions among themselves than with players on the other teams. We consider using two Co-LSTMs to model the inter-related dynamics among players on two teams, respectively. The new framework of H-LSTCM is shown in Figure~\ref{twogroup}. The main extension is that a concatenation operation and an LSTM layer are added on the top of the Co-LSTM layer. In this framework, each sub-group is modeled by a Co-LSTM, and then the outputs of two Co-LSTMs are concatenated into a sequence of representations, which are input into an LSTM layer. Likewise, Co-LSTSM+ (introduced in Section~\ref{cad}) is also modified in this way. In the B2, we model dynamics of one team by B2, and add a concatenation operation and an LSTM layer on the top of the LSTM layer. In the baseline B4, the outputs of one team of multiple Single-Person LSTMs are pooled into a sequence of representations. The representations of the two teams are concatenated into a long representation which is then fed into an LSTM layer. The recognition accuracy of H-LSTCM and all the baselines is shown in Table~\ref{volleyball}, where ``lpass", ``rpass", ``lset", ``rset", ``lspike", ``rspike", ``lwin" and ``rwin" denote left\_pass, right\_pass, left\_set,  right\_set, left\_spike, right\_spike, left\_winpoint and right\_winpoint, respectively. H-LSTCM achieves the best performance over all baseline methods. It is noted that Co-LSTSM+ and B4 are comparable. These results illustrate that Co-LSTSM+ cannot learn concurrently inter-related representations between multiple persons well, when a complex pattern of group activity exists. The confusion matrix of H-LSTCM is shown in Appendix C of the supplementary material.

		\begin{figure}[!t]
		\centering
		\includegraphics[scale=0.32]{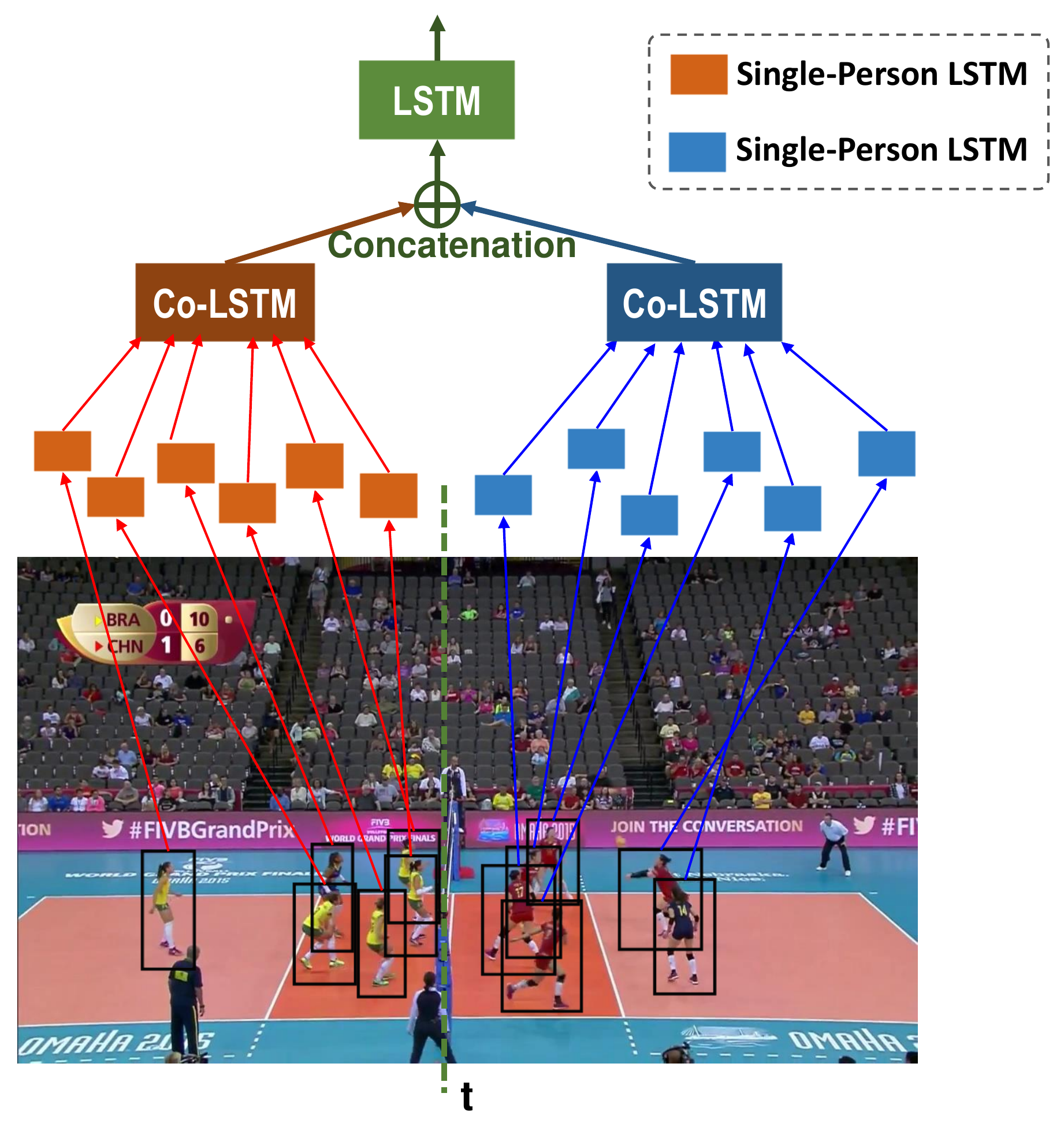}
		\vspace{-3mm}
		\caption{{Framework of H-LSTCM on the Volleyball activity with two sub-groups of persons. A concatenation operation and a LSTM layer is added on the top of Co-LSTM.}}\label{twogroup}
		\vspace{-3mm}
		
	\end{figure}	

\begin{table}[!t]
	\centering
	\caption{{ Recognition accuracy (\%) on Volleyball dataset.}}
	\vspace{-1mm}
	{\scriptsize
		\label{volleyball}
		\begin{tabular}{lcccccccccc}
			\hline
			\hspace{-0.5em}Method\hspace{-0.5em} & \hspace{-0.5em}{lpass}\hspace{-0.5em} & \hspace{-0.5em}{rpass}\hspace{-0.5em} & \hspace{-0.5em}{ lset}\hspace{-0.5em} & \hspace{-0.5em}{ rset}\hspace{-0.5em} & \hspace{-0.5em}{ lspike}\hspace{-0.5em} & \hspace{-0.5em}{ rspike}\hspace{-0.5em} & \hspace{-0.5em}{lwin}\hspace{-0.5em} & \hspace{-0.5em}{rwin}\hspace{-0.5em} &  \hspace{-0.5em}Averae\hspace{-0.7em}\\
			\hline
			\hspace{-0.5em}Ibrahim~\!\em{et al.}\cite{ibrahim2015hierarchical}\hspace{-0.5em} & \hspace{-0.5em}77.9\hspace{-0.5em}          & \hspace{-0.5em}81.4\hspace{-0.5em}          & \hspace{-0.5em}84.5\hspace{-0.5em}          & \hspace{-0.5em}68.8\hspace{-0.5em}          & \hspace{-0.5em}89.4\hspace{-0.5em}          & \hspace{-0.5em}85.6\hspace{-0.5em}          & \hspace{-0.5em}88.2\hspace{-0.5em}          & \hspace{-0.5em}87.4\hspace{-0.5em}            & \hspace{-0.5em}82.9\hspace{-0.5em} \\
			\hspace{-0.5em}Shu \emph{et al.} \cite{shu2016cern}\hspace{-0.5em}                 &\hspace{-0.5em} - \hspace{-0.5em}            & -             & -             & -             & -             & -             & -             & -               & \hspace{-0.5em}83.6\hspace{-0.7em} \\
			\hspace{-0.5em}Li \em{et al.} \cite{li2017sbgar}\hspace{-0.5em}                  & \hspace{-0.5em}55.8\hspace{-0.5em}          & \hspace{-0.5em}69.1 \hspace{-0.5em}         & \hspace{-0.5em}67.3\hspace{-0.5em}          & \hspace{-0.5em}52.1\hspace{-0.5em}          & \hspace{-0.5em}82.1\hspace{-0.5em}          & \hspace{-0.5em}79.2\hspace{-0.5em}          & -             & -               & \hspace{-0.5em}67.6\hspace{-0.7em} \\
			\hspace{-0.5em}Biswas~\!\em{et al.}\cite{biswas2018structural}\hspace{-0.5em}                 & -             & -             & -             & -             & -             & -             & -             & -               & \hspace{-0.5em}83.0\hspace{-0.5em} \\
			\hline
			\hspace{-0.5em}B1 & \hspace{-0.5em}~62.8\hspace{-0.5em}&	\hspace{-0.5em}62.1\hspace{-0.5em}&	\hspace{-0.5em}71.4\hspace{-0.5em}	&\hspace{-0.5em}58.7\hspace{-0.5em}	&\hspace{-0.5em}65.1\hspace{-0.5em}&	\hspace{-0.5em}76.5\hspace{-0.5em}&	\hspace{-0.5em}63.7\hspace{-0.5em}&	\hspace{-0.5em}61.6\hspace{-0.5em}&	\hspace{-0.5em}65.2\hspace{-0.5em}
			\\
			\hspace{-0.5em}B2  & \hspace{-0.5em}~64.6\hspace{-0.5em}&	\hspace{-0.5em}66.5\hspace{-0.5em}&	\hspace{-0.5em}76.5\hspace{-0.5em}&	\hspace{-0.5em}62.7\hspace{-0.5em}&	\hspace{-0.5em}77.7\hspace{-0.5em}&	\hspace{-0.5em}74.0\hspace{-0.5em}&	\hspace{-0.5em}70.6\hspace{-0.5em}&	\hspace{-0.5em}68.0\hspace{-0.5em}&	\hspace{-0.5em}70.1\hspace{-0.5em}
			\\
			\hspace{-0.5em}B3 & \hspace{-0.5em}~74.4\hspace{-0.5em}&	\hspace{-0.5em}77.3\hspace{-0.5em}&	\hspace{-0.5em}81.8\hspace{-0.5em}&	\hspace{-0.5em}69.7\hspace{-0.5em}&	\hspace{-0.5em}88.2\hspace{-0.5em}&	\hspace{-0.5em}83.7\hspace{-0.5em}&	\hspace{-0.5em}78.6\hspace{-0.5em}&	\hspace{-0.5em}78.0\hspace{-0.5em}&	\hspace{-0.5em}79.0\hspace{-0.5em}	
			\\
			\hspace{-0.5em}B4 & \hspace{-0.5em}~77.0\hspace{-0.5em}&	\hspace{-0.5em}80.9\hspace{-0.5em}&	\hspace{-0.5em}84.1\hspace{-0.5em}&	\hspace{-0.5em}68.3\hspace{-0.5em}&	\hspace{-0.5em}88.8\hspace{-0.5em}&	\hspace{-0.5em}85.3\hspace{-0.5em}&	\hspace{-0.5em}88.0\hspace{-0.5em}&	\hspace{-0.5em}87.7\hspace{-0.5em}&	\hspace{-0.5em}82.5\hspace{-0.5em}	
			\\
			\hspace{-0.5em}{Co-LSTSM+} & \hspace{-0.5em}~81.3\hspace{-0.5em}& 	\hspace{-0.5em}79.5\hspace{-0.5em}&	\hspace{-0.5em}85.1\hspace{-0.5em}&	\hspace{-0.5em}70.7\hspace{-0.5em}&	\hspace{-0.5em}88.8\hspace{-0.5em}&	\hspace{-0.5em}85.5\hspace{-0.5em}&	\hspace{-0.5em}88.7\hspace{-0.5em}&	\hspace{-0.5em}86.9\hspace{-0.5em}&	\hspace{-0.5em}83.3\hspace{-0.5em}
			\\
			\hspace{-0.5em}{H-LSTCM} & \hspace{-0.5em}~83.9\hspace{-0.5em}&	\hspace{-0.5em}88.1\hspace{-0.5em}&	\hspace{-0.5em}90.3\hspace{-0.5em}&	\hspace{-0.5em}80.4\hspace{-0.5em}&	\hspace{-0.5em}93.4\hspace{-0.5em}&	\hspace{-0.5em}89.8\hspace{-0.5em}&	\hspace{-0.5em}88.7\hspace{-0.5em}&	\hspace{-0.5em}92.4\hspace{-0.5em}&	\hspace{-0.5em}88.4\hspace{-0.5em}
			\\
			\hline
		\end{tabular}
	}
\end{table}

		{\bf Comparison with state-of-the-art methods.}  The results  of the proposed H-LSTCM and other related methods are also shown in Table~\ref{volleyball}.  H-LSTCM achieves the higher recognition accuracy than the state-of-the-art methods, including Ibrahim {\em et al.}~\cite{ibrahim2015hierarchical}, Shu {\em et al.}~\cite{shu2016cern}, Li {\em et al.}~\cite{li2017sbgar}, and Biswas {\em et al.}~\cite{biswas2018structural}. In particular, H-LSTCM with an accuracy of 88.4\% achieves approximately 5\% improvement compared with Shu et al. with an accuracy of 83.6\%. This demonstrates that H-LSTCM is effective in modeling complex collective activity among a sub-group of persons. Finally, we present some recognition results of H-LSTCM in Figure~\ref{fig_results}(d).
	}


	\subsection{Evaluation on Human Interaction Prediction}
	We also evaluate H-LSTCM on human interaction prediction. In contrast to human interaction recognition, human interaction prediction is defined as recognizing an ongoing interaction activity before the interaction is completely executed~\cite{ke2016spatial,ryoo2011human}. Due to the large variations in appearance and the evolution of scenes, human interaction prediction is a challenging task. Following the experimental setting in~\cite{ke2016spatial,kong2014max}, a testing video clip is divided into 10 incomplete action executions by using 10 observation ratios (i.e., from 0 to 1 with a step size of 0.1), which represent the increasing amount of sequential data with time. For example, given a testing video clip of length $T$, an observation ratio of $0.3$ denotes that the accuracy is tested with the first $0.3\times T$ frames. When the observation ratio is $1$, namely the entire video clip is used, H-LSTCM acts as a human interaction recognition model.

		\begin{figure}[!t]
		
		\centering
		\subfigure[On the BIT dataset.]{
			\includegraphics[scale=0.365]{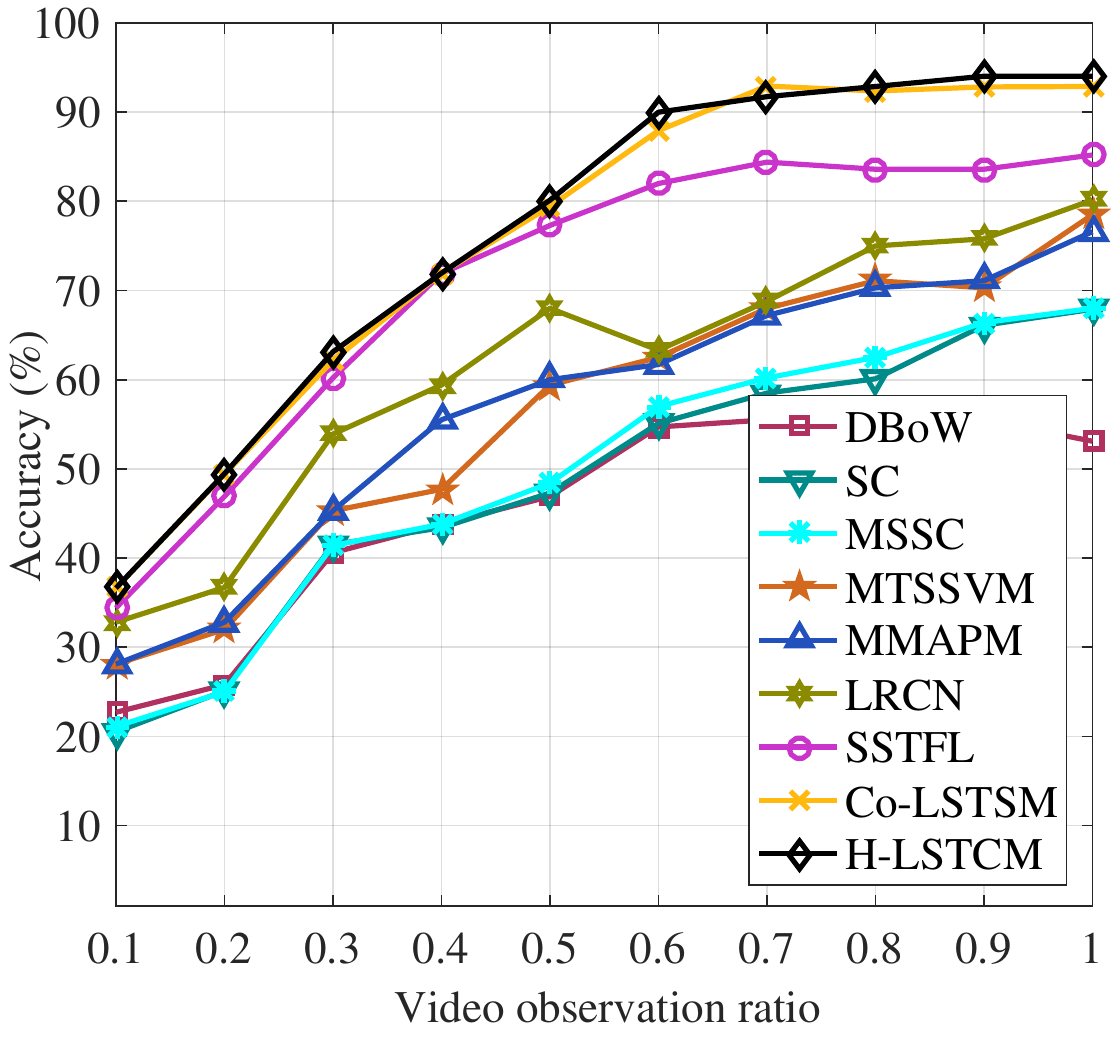}
			\label{fig_prediction_BIT}
		}
		\subfigure[On the UT dataset.]{
			\includegraphics[scale=0.365]{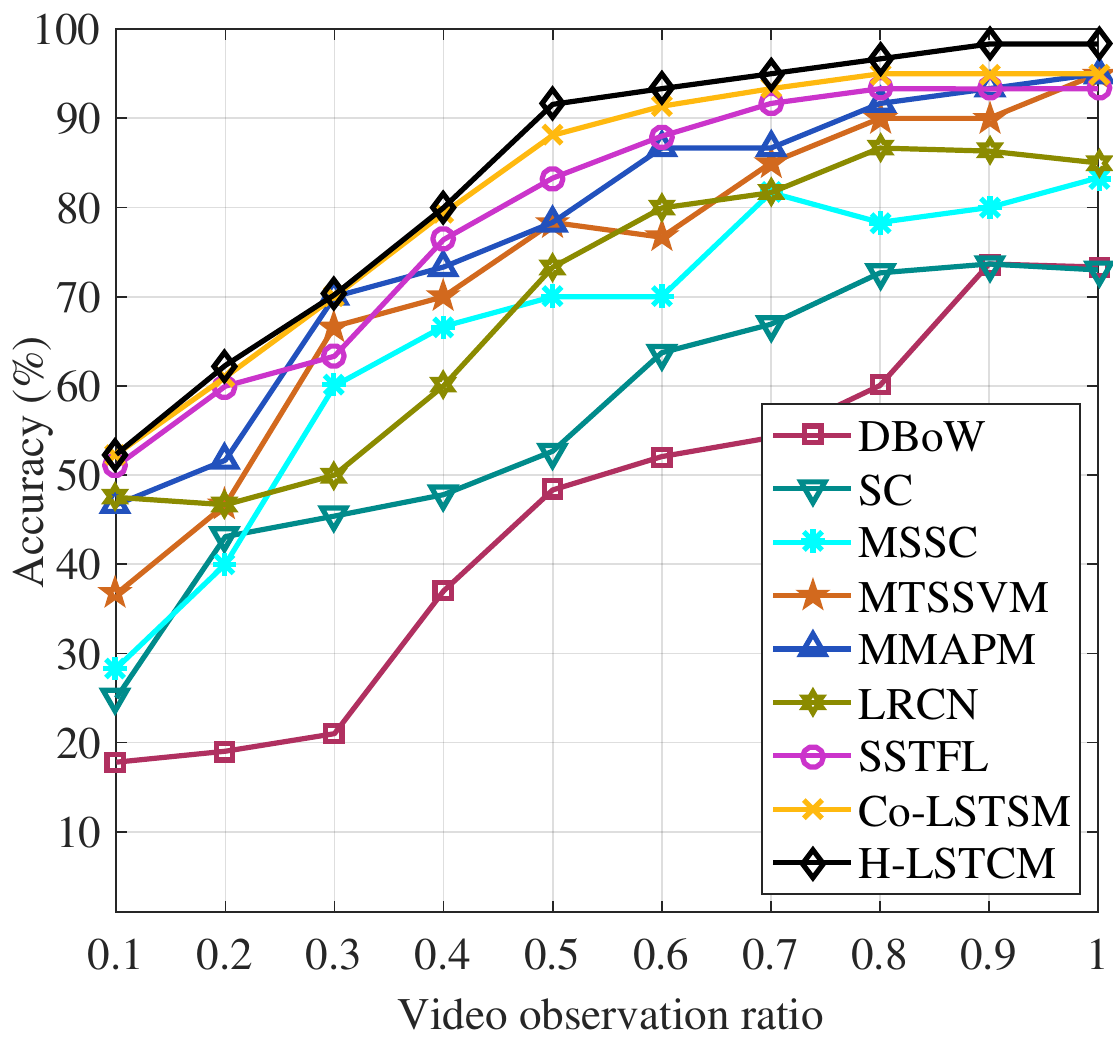} 
			\label{fig_prediction_UT}
		}
		\vspace{-3mm}
		\caption{Comparisons of human interaction prediction on BIT and UT.}
		\label{fig_prediction}
		\vspace{-3mm}
	\end{figure}
	
	The baselines include Dynamic
	Bag-of-Words (DBoW)~\cite{ryoo2011human}, Sparse Coding (SC)~\cite{cao2013recognize}, Sparse Coding with Mixture
	of training video Segments (MSSC)~\cite{cao2013recognize}, Multiple Temporal Scales based on SVM (MTSSVM)~\cite{kong2014a}, Max-Margin Action Prediction Machine (MMAPM)~\cite{kong2014max}, Long-term Recurrent Convolutional Networks (LRCN)~\cite{donahue2015long}, Spatial-Structural-Temporal Feature Learning (SSTFL)~\cite{ke2016spatial} and our preliminary Co-LSTSM ~\cite{shu2017concurrence}. The results of all the methods on the BIT and UT datasets with different observation ratios are listed
	in Figure~\ref{fig_prediction_BIT} and Figure~\ref{fig_prediction_UT}, respectively. Overall, H-LSTCM and Co-LSTSM outperforms all the baselines. All the interactions in the BIT dataset are the two persons’ interactions with simple background, and Co-LSTSM is proposed to learn the dynamic inter-related representation between two persons. Thus, the performance of H-LSTCM is comparable to Co-LSTSM for the problem of two persons’ interaction predication on the BIT dataset. Specifically, we can observe that: 1) the improvements of H-LSTCM and Co-LSTSM on BIT are more significant when the observation ratio is $0.6$; 2) the accuracy of H-LSTCM becomes stable on both BID and UT when the observation ratio is approximately $0.8$, which illustrates the end of close interaction is ending; and 3) since H-LSTSM and Co-LSTSM can accumulate the temporal interacting information, their accuracy monotonously increases with increasing video observation ratio.

	\vspace{-0mm}
	\section{Conclusions}
	\label{C}

	In this work, on human interaction recognition, we propose a novel Hierarchical Concurrent Long Short-Term Concurrent Memory (H-LSTCM) to learn the dynamic inter-related representation among all persons from the static singe-person features in a hierarchical way. Specifically, for each person, we first feed her/his static single-person features into a Single-Person LSTM to learn the single-person dynamic. Afterwards, the outputs of all Single-Person LSTMs unit are fed into a novel Concurrent LSTM (Co-LSTM) unit, which mainly consists of multiple sub-memory units and a new co-memory cell. In the Co-LSTM unit, each sub-memory unit stores individual motion information, while a concurrent LSTM (Long Short-Term Memory) unit selectively integrates and stores the inter-related motion information among multiple interacting persons from multiple sub-memory units via a new co-memory cell. The proposed method is evaluated on four public datasets and yields promising improvements over the state-of-the-art methods. 

	
	
	\vspace{-1mm}
	\bibliographystyle{IEEEtran}
	\bibliography{IEEEabrv,egbib_new}

\end{document}